\documentclass[nohyperref]{article}

\usepackage{microtype}
\usepackage{graphicx}
\usepackage{subfigure}
\usepackage{booktabs} 

\usepackage{hyperref}

\usepackage[accepted]{icml2022}

\usepackage{amsmath}
\usepackage{amssymb}
\usepackage{mathtools}
\usepackage{amsthm}

\usepackage{multirow}
\usepackage{hhline}

\usepackage[all=normal,lists=tight,floats=tight,paragraphs=tight]{savetrees}

\usepackage[capitalize,noabbrev]{cleveref}

\theoremstyle{plain}

\theoremstyle{definition}

\theoremstyle{remark}

\usepackage[textsize=tiny]{todonotes}

\icmltitlerunning{Improved Robustness Against Adaptive Attacks With Ensembles and Error-Correcting Output Codes}

\begin{document}

\twocolumn[
\icmltitle{Improved Robustness Against Adaptive Attacks With\\ Ensembles and Error-Correcting Output Codes}



\icmlsetsymbol{equal}{*}

\begin{icmlauthorlist}
\icmlauthor{Thomas Philippon}{ulaval}
\icmlauthor{Christian Gagné}{ulaval,CCAI}
\end{icmlauthorlist}

\icmlaffiliation{ulaval}{Institute Intelligence and Data (IID), Université Laval, Quebec, Canada}
\icmlaffiliation{CCAI}{Canada CIFAR AI Chair, Mila}

\icmlcorrespondingauthor{Thomas Philippon}{thomas.philippon.1@ulaval.ca}

\icmlkeywords{Adversarial Machine Learning, Adversarial Defense, Error-Correcting Output Codes, Ensemble Learning, Neural Network Ensemble, Adaptive Adversarial Attack}

\vskip 0.3in
]



\printAffiliationsAndNotice{}  

\begin{abstract}
Neural network ensembles have been studied extensively in the context of adversarial robustness and most ensemble-based approaches remain vulnerable to adaptive attacks. In this paper, we investigate the robustness of Error-Correcting Output Codes (ECOC) ensembles through architectural improvements and ensemble diversity promotion. We perform a comprehensive robustness assessment against adaptive attacks and investigate the relationship between ensemble diversity and robustness. Our results demonstrate the benefits of ECOC ensembles for adversarial robustness compared to regular ensembles of convolutional neural networks (CNNs) and show why the robustness of previous implementations is limited. We also propose an adversarial training method specific to ECOC ensembles that allows to further improve robustness to adaptive attacks. 
\end{abstract}

\section{Introduction}
\label{submission}

Deep neural networks are known to be vulnerable to adversarial perturbations \cite{szegedy2013intriguing, goodfellow2014}. To increase robustness to such attacks, various ensemble-based defenses were adopted \cite{pang2019improving, sen2020empir, DBLP:journals/corr/AbbasiG17, vermaECOC}. Nevertheless, there is room for improvement given that most of these defenses were shown to be vulnerable to adaptive attacks (i.e., adversarial attacks adapted to a specific defense mechanism) \cite{tramer2020adaptive, he2017adversarial}. 

Error-Correcting Output Codes (ECOC) ensembles are inspired from information theory and were introduced by \citet{dietterich1994solving} for solving multi-class learning problems. More recently, they were used to defend against adversarial attacks \cite{vermaECOC}, but were later shown vulnerable to adaptive attacks by  \citet{tramer2020adaptive}. In this work, we revisit ECOC ensembles in the context of adversarial robustness through architectural improvements and perform a comprehensive experimental robustness assessment against white-box adaptive attacks. We also explore the relationship between ensemble diversity and robustness for ECOC ensembles of different architectures. Finally, since the best adversarial defenses are currently based on adversarial training \cite{bai2021recent}, we investigate the combination of ECOC ensembles and adversarial training.

Our main contributions consist of:
\vspace{-0.8em}
\begin{itemize}
\item Providing an extensive robustness assessment for ECOC ensembles of different architectures to demonstrate their robustness to adaptive attacks when they are composed of fully independent base classifiers. Our results also demonstrate the superior robustness of the proposed ECOC ensembles with fully independent base classifiers compared to regular ensembles of CNNs with soft-voting.

\item Investigating ensemble diversity among the base classifiers for ECOC ensembles of various architectures to show empirically the relationship between diversity and robustness. 

\item Demonstrating that adversarial training can be combined effectively with ECOC ensembles when the base classifiers forming ECOC ensembles are trained independently, with their own adversarial perturbations, instead of trained with the same perturbations generated on the entire ensemble. 
\end{itemize}

\section{Background}
\subsection{Error-Correcting Output Codes}

In the approach proposed by \citet{dietterich1991error}, the $M$ classes of multi-class learning problems are represented by unique $N$-bit class codewords (i.e. binary strings) stored in $M \times N$ codeword matrices (Tab.~\ref{tab:codewordsexample}). For solving such problems, $N$-bit ECOC ensembles are used. They are composed of $N$ binary classifiers, acting as base classifiers, solving different binary classification tasks represented in the columns of the $M \times N$ matrix, as shown in Table~\ref{tab:codewordsexample}. During inference, the output values from the $N$ binary classifiers are concatenated to form a predicted $N$-bit codeword for each input image. Then, to obtain the inferred class from a predicted codeword, the nearest class codeword is determined with the Hamming distance, which reports the number of distinct bits between two binary codes.

\begin{table}[t]
\caption{Example of a codeword matrix for $M=5$ classes and $N=10$ bit codewords. This matrix is used for ECOC ensembles composed of 10 binary classifiers solving distinct binary classification problems represented by the 10 rows of the matrix. A +1/-1 encoding is used with this matrix to simplify computations.}
\label{tab:codewordsexample}
\begin{center}
\begin{small}
\setlength{\tabcolsep}{3pt}
\begin{sc}
\begin{tabular}{ c|rrrrrrrrrr }
\toprule
  \multirow{2}{*}{Class} & \multicolumn{10}{c}{Binary classifiers} \\
 & $f_1$ &$f_2 $ &$f_3$  &$f_4$ &$f_5$ &$f_6$ &$f_7$ &$f_8$ &$f_9$ &$f_{10}$ \\
 \midrule
$C_1$ & -1 & -1 & -1 & +1 & +1 & -1 & -1 & -1 & -1 & +1\\
$C_2$ & +1 & -1 & -1 & -1 & -1 & -1 & +1 & -1 & +1 & -1\\
$C_3$ & +1 & -1 & +1 & +1 & +1 & -1 & +1 & +1 & +1 & -1\\
$C_4$ & -1 & -1 & +1 & -1 & +1 & +1 & -1 & +1 & -1 & -1\\
$C_5$ & -1 & +1 & -1 & +1 & -1 & -1 & +1 & +1 & +1 & -1\\
\bottomrule
\end{tabular}
\end{sc}
\end{small}
\end{center}
\vskip -0.3in
\end{table}

According to \citet{dietterich1994solving}, the error correcting capabilities inspired by coding theory is the main benefit of ECOC ensembles. Indeed, if the minimum Hamming distance between any pair of class codewords in the codeword matrix is $\theta$, the ensemble will be robust to $\lfloor\frac{\theta-1}{2}\rfloor$ binary classifiers making wrong predictions \cite{dietterich1994solving}. Another benefit of ECOC ensembles is the diversity among their base classifiers. As opposed to traditional ensembles with base classifiers trained on a similar task using one-hot encodings, the $N$ binary classifiers of ECOC ensembles are trained on unique tasks, represented by different splits of the class set in two groups (i.e, columns of the codeword matrix). Diversity can thus be promoted by maximizing the column separation in the codeword matrix. Indeed, having two identical columns or the complement of a column equal to another column leads to base classifiers trained on similar splits of the class set, and thus classifiers more likely to make similar mistakes. 

\subsection{ECOC as an Adversarial Defense Mechanism}
Motivated by their error correcting capabilities and the diversity among their base classifiers, ECOC ensembles have been recently proposed as an adversarial defense mechanism \cite{vermaECOC}. It is argued that since the binary classifiers of ECOC ensembles learn different tasks and are thus more diverse, it is more difficult for an attacker to generate perturbations that can fool all of them at once. Therefore, since ECOC ensembles can be robust to some of their binary classifiers making wrong predictions, they will be robust to adversarial attacks as long as no more than $\lfloor\frac{\theta-1}{2}\rfloor$ binary classifiers are fooled. 

The ECOC ensembles of \citet{vermaECOC} are composed of $N$ CNN binary classifiers sharing several layers among them to extract common features -- the base classifiers are organized in groups of four models sharing the same feature extractor layers. 

The robustness assessment in \citet{vermaECOC} showed promising results. The best ECOC model trained on CIFAR10 had an accuracy rate of 57.4\% against strong adversarial attacks bounded by an $\ell_{\infty}$ norm of 0.031. However, \citet{tramer2020adaptive} later showed the claimed accuracy of 57.4\% could be dropped to less than 5\% using adaptive attacks bounded by the same norm of 0.031. This was achieved by computing adversarial perturbations directly on the output bits, bypassing consecutive numerically unstable operations at the output of the ECOC ensembles such as $\tanh$, $\log$ and $\mathrm{softmax}$. These operations were causing gradient masking, preventing regular adversarial attacks from generating proper adversarial perturbations for ECOC ensembles \citep{tramer2020adaptive}.

\subsection{ECOC and Adversarial Training}

Adversarial training was first introduced by \citet{szegedy2013intriguing} and further developed by \citet{goodfellow2014, madry2017towards,zhang2019theoretically}. It consists of augmenting or replacing the training data with adversarial perturbations generated from some training data instances to promote adversarial robustness. \citet{song2021error} investigated the combination of ECOC and adversarial training. They reported an increase in robustness when training all the binary classifiers with the same perturbations generated on the whole ensemble using Projected Gradient Descent (PGD) with the cross-entropy loss \cite{madry2017towards}.

\section{Methodology}


In this work, we revisit ECOC ensembles as an adversarial defense by investigating different architectures so as to increase robustness to adaptive attacks. In the following, we propose an approach to ECOC ensembles designed to increase resilience to adversarial attacks in comparison of that of previous proposals\footnote{The source code is available at: \url{https://github.com/thomasp05/Improved-Robustness-with-ECOC}}.

\subsection{ECOC Architecture}

In line with the original ECOC architecture of \citet{dietterich1994solving}, our ECOC models are ensembles of $N$ fully independent CNN binary classifiers. Even though it is a simple configuration, it is the first time, to the best of our knowledge, that such an architecture is investigated in the context of adversarial robustness for ECOC ensembles. Other ECOC architectures proposed for adversarial robustness share some parts of the binary classifiers (i.e. feature extractor layers) as a matter of efficiency \cite{vermaECOC,song2021error}. We believe that using independent CNN binary classifiers to construct the ECOC ensembles will increase robustness by promoting the learning of more diverse features, making the binary classifiers more challenging to fool simultaneously with adversarial attacks. In the remainder of the paper, we denote ECOC ensembles with such an architecture as $\text{ECOC}_{N,1}$. In the experiments, we use binary classifiers composed of 11 convolution layers followed by a fully connected layer outputting a single bit. This architecture is inspired by that of \cite{vermaECOC} -- see Sec.~\ref{app:architecture} of the supplementary material for more details. 

We also implemented ECOC ensembles with CNN binary classifiers sharing feature extractor layers, where $\frac{N}{K}$ shared feature extractors are connected to $K$ parallel heads, each outputting a single bit. Each feature extractor is thus shared by $K$ binary classifiers. Inspired by \citet{vermaECOC}, each feature extractor is composed of 9 convolution layers connected to $K$ parallel heads composed of 2 convolution layers and a fully connected layer with a single bit as output -- see again Sec.~\ref{app:architecture} for more details. ECOC ensembles with such an architecture are denoted as $\text{ECOC}_{N,K}$ herein.

The binary classifiers for both architectures are trained using a hinge loss computed on each bit independently:
\begin{equation}
\ell_{\text{hinge}}(n) = \max(1-z_n\,a_{i,n}, 0),
\label{eq:hinge loss}
\end{equation}
where $z_n\in\mathbb{R}$ represents the predicted value at index $n\in\{1,\ldots,N\}$ and $a_{i,n}\in\{-1,1\}$ is the corresponding bit in the ground truth class codeword $C_i$.

In the original ECOC architecture \cite{dietterich1994solving}, the ensemble predictions rely on the Hamming distance between the predicted and the class codewords. In our case, this non-differentiable approach is problematic since white-box adversarial attacks rely on output logits gradients to generate perturbations. We consider a differentiable decoder function to map predicted codewords to class probabilities while making sure to avoid inducing spurious robustness from numerically unstable operations \cite{tramer2020adaptive}. With our decoder function, each binary classifier output value $z_j\in\mathbb{R}$ is passed through a $\tanh$ function to scale it back to the $[-1,1]$ range. Then, each of these outputs is multiplied with the class codewords $\mathbf{a}_m\in\{-1,1\}^N$, where $m\in\{1,\ldots,M\}$, leading to:
\begin{equation}
h_m = \sum_{j=1}^N a_{m,j} \tanh(z_j).
\label{eq:decoder}
\end{equation}
The predicted class $y$ is determined according to the maximum $h_m$ output:
\begin{equation}
y = {\arg\max}_{i=1}^M h_i.
\end{equation}
A softmax function can be applied to the vector $\mathbf{h}=[h_1 \cdots h_M]^\top$ to obtain class probabilities, useful to generate adversarial perturbations with white-box attacks. 

\subsection{Code Design}

Codeword design is central to ECOC ensembles \cite{dietterich1994solving}. It determines how many individual errors from the binary classifiers can be handled without making ensemble misclassifications and how diverse the binary classification tasks are. In our work, we rely on a guided random generation of 16-, 32- and 64-bit codewords. The use of randomly generated codewords is well motivated since it has been demonstrated to perform well \cite{dietterich1991error, james1998error, ahmed2021deep}. Other works that focus on codeword design for ECOC ensembles include \citet{youn2021construction} and \citet{gupta2021integer}.

Our approach aims at achieving two properties when randomly generating a $M \times N$ codeword matrix $\mathbf{A}$. First, we ensure that the Hamming distance between each codeword pair (rows $\mathbf{a}_i$ of $\mathbf{A}$) is at least $\theta_\text{minham}$:
\begin{equation}
\mathrm{d}_\text{ham}(\mathbf{a}_i, \mathbf{a}_j) \geq \theta_\text{minham}, \quad \forall_{\substack{i,j\in\{1,\ldots,M\}\\ i\neq j}}.
\label{eq:minham}
\end{equation}
This property allows the ensemble to correct up to $\lfloor\frac{\theta_\text{minham}-1}{2}\rfloor$ misclassifications from the $N$ binary classifiers \cite{dietterich1994solving}.

The second property is achieved by using two conditions: 1) the minimum Hamming distance between each classifier decision profile pair (columns $\mathbf{a}_{\cdot,j}$ of $\mathbf{A}$) should be at least $\theta_\text{div}$, and 2) the minimum Hamming distance between each pair composed of the complement (inverse) of a classifier decision profile and a standard decision profile should be at least $\theta_\text{cdiv}$. The use of the complement in the decision profiles allows the trivial cases to be handled where different classifiers make exact opposite decisions, which is not diverse (i.e., inverting the decision profile of one classifier leads to a very similar model). The purpose of the second property is to promote diversity between the decision profiles of the binary classifiers. The two conditions are captured in the following equations:
\begin{align}
\mathrm{d}_\text{ham}(\mathbf{a}_{\cdot,i}, \mathbf{a}_{\cdot,j}) & \geq \theta_\text{div}, \quad \forall_{\substack{i,j\in\{1,\ldots,N\}\\ i\neq j}},\label{eq:div}\\
\mathrm{d}_\text{ham}(\mathbf{\bar{a}}_{\cdot,i}, \mathbf{a}_{\cdot,j}) & \geq \theta_\text{cdiv}, \quad \forall_{\substack{i,j\in\{1,\ldots,N\}\\ i\neq j}},\label{eq:cdiv}
\end{align}
where $\bar{\mathbf{a}}_{\cdot,i}$ is the complement of the decision profile $\mathbf{a}_{\cdot,i}$.

Our random codeword generation proceeds by generating $M$ random rows sequentially. Every new row must respect Eq.~\ref{eq:minham} or else another random row is generated until it does respect Eq.~\ref{eq:minham}. When $M$ rows are generated to form the codeword matrix $\mathbf{A}$, the conditions given as Eq.~\ref{eq:div} and Eq.~\ref{eq:cdiv} are verified. Randomly selected rows of $\mathbf{A}$ are then replaced with new rows respecting Eq.~\ref{eq:minham} one at a time until the conditions of Eq.~\ref{eq:div} and Eq.~\ref{eq:cdiv} are respected. The numerical values of the parameters used to generate the 16-, 32- and 64-bit codeword matrices are ($\theta_\text{minham}=8$, $\theta_\text{div}=3$, $\theta_\text{cdiv}=3$), 
($\theta_\text{minham}=16$, $\theta_\text{div}=2$, $\theta_\text{cdiv}=1$), and ($\theta_\text{minham}=32$, $\theta_\text{div}=1$, $\theta_\text{cdiv}=1$), respectively.

\subsection{Adversarial Training}

Many of the most effective defenses rely on adversarial training \cite{madry2017towards, zhang2019theoretically}. For ECOC in particular, some have proposed to augment the training data at the ensemble level, with perturbations generated on the output logits of the decoder \cite{song2021error}. We argue that this is likely not the best way to perform adversarial training with ECOC ensembles. The perturbations generated at the ensemble level cannot fool all of the binary classifiers in the ensemble. Indeed, it is required to fool only $\lfloor\frac{\theta_\text{minham}-1}{2}+1\rfloor$ binary classifiers in order to fool an ECOC ensemble. Consequently, some classifiers are trained with unspecific perturbations which could prevent them from learning robust features through adversarial training.

In our case, we augment the training data with perturbations generated on each binary classifier forming the ECOC ensemble. To do so, we modified the original PGD attack \cite{madry2017towards} to generate perturbations using the hinge loss on a single binary classifier output. At every epoch, specific adversarial perturbations are generated for each of the $N$ binary classifiers. The images forming the training batch of each classifier are then replaced with their specific perturbed versions, with the rest of the training steps remaining the same. This approach will allow each binary classifier to be trained with specific and relevant perturbations. We believe this will help promote the learning of robust features and increase the robustness of ECOC ensembles. The perturbations should also require fewer iterations for PGD. We validated our proposal by means of a comparison of both adversarial training with data perturbations generated for individual classifiers (henceforth IndAdvT) and adversarial training with perturbations generated on the whole ensemble as in \citet{song2021error} (henceforth RegAdvT).

\section{Experiments}

Three vanilla CNN models are used as baselines in our experiments for comparison with the ECOC ensembles. The first baseline, named $\text{SIMPLE}$, is similar to a single independent network used in our $\text{ECOC}_{N,1}$ ensembles. It consists of a regular CNN composed of 11 convolution layers followed by a fully connected layer outputting 10 neurons trained with the cross-entropy loss. The two other baselines are $\text{ENSEMBLE}_{6}$ and $\text{ENSEMBLE}_{16}$, which are ensembles of 6 and 16 independent $\text{SIMPLE}$ models. They are combined with soft-voting, that is the output of the individual models are passed through the softmax function and summed such that the final prediction is the class with the largest sum probability -- see Sec.~\ref{app:architecture} for more details. We used the CIFAR10 and Fashion-MNIST datasets. All models are trained with the Adam optimizer and a batch size of 100. The ECOC models for CIFAR10 are trained for 900 epochs with a learning rate of $10^{-4}$ and for 100 additional epochs with a learning rate of $10^{-5}$. The three baseline models for CIFAR10 are trained for 400 epochs with a learning rate of $10^{-3}$. For Fashion-MNIST, the ECOC models are trained for 60 epochs with a learning rate of $10^{-5}$ and the baseline models for 15 epochs with a learning rate of $10^{-4}$. These parameters were selected to optimize the performance and robustness of each architecture. 

\subsection{Robustness Assessment}

The results of the robustness assessment for the CIFAR10 and Fashion-MNIST models are shown in tables \ref{tab:adv attacks} and \ref{tab:adv attacks fmnist}. Robustness is assessed in a white-box setting against untargeted attacks. For models trained one both datasets, attacks are performed on 2000 randomly selected test images and the clean accuracy is computed on all 10000 test images. The regular attacks used in tables \ref{tab:adv attacks} and \ref{tab:adv attacks fmnist} are Fast Gradient Sign Method (FGSM) \cite{goodfellow2014}, Projected Gradient Descent (PGD) \cite{madry2017towards} and Carlini and Wagner ($\text{C\&W}_{\ell_2}$) \cite{carlini2017towards}. FGSM and PGD are $\ell_{\infty}$ attacks while $\text{C\&W}_{\ell_2}$ is an $\ell_{2}$ attack. The adaptive attacks, inspired by \citet{tramer2020adaptive} are labeled $\text{PGD}^{es}$, $\text{PGD}^{es+}$, $\text{PGD}_{cw}$, $\text{PGD}_{cw}^{+}$, $\text{PGD}_{cw}^{es}$, $\text{PGD}_{cw}^{es+}$ and $\text{C\&W}_{\ell_2}^{+}$. The $cw$ tag refers to a version of the PGD attack where the cross entropy loss is replaced with the margin loss introduced by \citet{carlini2017towards}. The $es$ tag indicates that a form of early stopping is used. Instead of systematically returning the perturbations at the last iteration of PGD, the last successful perturbations are returned. The + tag indicates that the model under attack was modified to remove operations that could mask the gradient. More specifically, the $\tanh$ operation of Eq.~\ref{eq:decoder} for ECOC and the softmax operation of the soft-voting scheme for $\text{ENSEMBLE}_{6}$ and $\text{ENSEMBLE}_{16}$. This modification does not apply to the $\text{SIMPLE}$ model. The attacks used are from the CleverHans library V4.0.0 \cite{cleverhans}. See Sec.~\ref{app:adv attacks} of the supplementary material for more details. 

The attacks used to generate the results in Table~\ref{tab:adv attacks} are bounded by an $\ell_{\infty}$ norm of 0.031 or an $\ell_{2}$ norm of 1.0. For Table~\ref{tab:adv attacks fmnist}, the bounds used are 0.06 for the $\ell_{\infty}$ norm and 1.5 for the $\ell_{2}$ norm. Attacks generated with PGD rely on 200 iterations and a step size of $\frac{\epsilon}{3}$, where $\epsilon$ is the $\ell_{\infty}$ bound used. For $\text{C\&W}_{\ell_2}$ attacks, we use a learning rate of $5\times 10^{-3}$, 1000 iterations, a confidence of 0 and 5 binary steps.

\begin{table*}[t]
\caption{Accuracy of models trained on CIFAR10 against images perturbed by different adversarial attacks. The accuracy for the most successful $\ell_{\infty}$ and $\ell_{2}$ attacks for each model are in bold.}
\label{tab:adv attacks}
\vskip 0.1in
\begin{center}
\begin{small}
\setlength{\tabcolsep}{4pt}
\begin{sc}
 \begin{tabular}{l|c|cccccccc|cc}
    \toprule
    Model  & Clean     & FGSM    &  PGD & $\text{PGD}^{es}$ & $\text{PGD}^{es+}$ & $\text{PGD}_{cw}$ & $\text{PGD}_{cw}^{+}$ & $\text{PGD}_{cw}^{es}$ & $\text{PGD}_{cw}^{es+}$ & $\text{C\&W}_{\ell_2}$ & $\text{C\&W}_{\ell_2}^{+}$ \\
    \midrule
    $\text{SIMPLE}$        &  84.4  &  20.2   &  0.7 & 0.7 & - &  \textbf{0.0}  &  -      &   \textbf{0.0}   & -  &   \textbf{0.0}    &   -   \\
    $\text{ENSEMBLE}_{6}$  &  89.9  &  49.1   &  0.5 & 0.4 & 21.8 &  0.6  &  \textbf{0.0}   &   0.5   & \textbf{0.0}  & 8.6  &  \textbf{0.0}   \\
    $\text{ENSEMBLE}_{16}$ &  90.9  &  45.3   &  0.4 & 0.4 & 61.3 &  0.4  &  0.1  &   0.4   & \textbf{0.0} & 4.2  &  \textbf{0.1}   \\\midrule
    $\text{ECOC}_{16,4}$   &  88.5  &  41.2   &  8.4 & \textbf{1.4}  &  12.1 &  8.7  &  7.5  & 2.0 & 1.4 & 7.2   &  \textbf{4.4}  \\
    $\text{ECOC}_{16,2}$   &  87.5  &  57.2  &  18.6 & \textbf{7.1} & 11.8 &  19.3 &  22.2  & 8.8 & 9.4  & 16.5  &  \textbf{12.2}  \\
    $\text{ECOC}_{16,1}$   &  85.8  &  67.9  &  41.1 & \textbf{26.4} &  27.7  &  41.8  &  43.7 & 26.6 & 27.5 & 30.5 &  \textbf{27.0}   \\\midrule
    $\text{ECOC}_{32,4}$   & 89.2   &  65.2  &  25.5 & 18.0 & 40.0 &  14.0 &  13.8  & \textbf{3.3} & 3.6 & 7.7  &  \textbf{4.9}  \\
    $\text{ECOC}_{32,2}$   &  88.6  &  67.7   &  28.9 & 20.1 & 34.6 &  24.1 &  24.9 & \textbf{11.7} & 12.8 & 12.8 &  \textbf{11.8}    \\
    $\text{ECOC}_{32,1}$   & 86.6  &  69.6  &  39.4 & 31.4 &  38.3 &  34.6 &  36.9  & \textbf{25.0} & 25.8  & 23.7  &  \textbf{20.9} \\\midrule
    $\text{ECOC}_{64,4}$   &  90.2   &  67.6  & 34.1 & 31.3 & 45.4 &  12.0  &  11.8  & \textbf{5.6} & 6.1 & 5.9  &  \textbf{3.7}  \\
    $\text{ECOC}_{64,2}$   &  89.2  &  68.5   &  35.6 & 32.1 & 39.4 &  19.7 &  19.8  & \textbf{12.0} & 13.2 & 11.9  & \textbf{9.7}  \\
    $\text{ECOC}_{64,1}$   &  87.5  &  69.5  &  37.6 & 34.0 & 38.6 &  30.8  & 31.5   & \textbf{23.6} & 25.3 & 19.4 & \textbf{16.3}  \\
    \bottomrule
  \end{tabular}
\end{sc}
\end{small}
\end{center}
\vskip -0.1in
\end{table*}

\begin{table*}[t]
\caption{Accuracy of models trained on Fashion-MNIST against images perturbed by different adversarial attacks. The accuracy for the most successful $\ell_{\infty}$ and $\ell_{2}$ attacks for each model are in bold.}
\label{tab:adv attacks fmnist}
\vskip 0.1in
\begin{center}
\begin{small}
\setlength{\tabcolsep}{4pt}
\begin{sc}
 \begin{tabular}{l|c|cccccccc|cc}
    \toprule
    Model  & Clean     & FGSM    &  $\text{PGD}$ & $\text{PGD}^{es}$ & $\text{PGD}^{es+}$ &  $\text{PGD}_{cw}$ & $\text{PGD}_{cw}^{+}$ & $\text{PGD}_{cw}^{es}$ & $\text{PGD}_{cw}^{es+}$ & $\text{C\&W}_{\ell_2}$ & $\text{C\&W}_{\ell_2}^{+}$ \\
    \midrule
    $\text{SIMPLE}$        &  85.9   &  13.3  &  2.6 & \textbf{2.3} &  - &   4.1 &  -  &  2.9  &  -  &  \textbf{0.7}  &  -  \\
    $\text{ENSEMBLE}_{6}$  &  87.5 &  24.1  &   5.9 & \textbf{5.7} & 8.6 & 6.8  & 8.5 &   6.7   &  8.1  &  3.5 & \textbf{2.2} \\
    $\text{ENSEMBLE}_{16}$ &  87.7  &  26.1  &  10.9 & \textbf{10.5} & 36.7 &  11.7  &  14.1  &   11.4   &  13.9  &  5.1  &  \textbf{3.9}  \\\midrule
    $\text{ECOC}_{16,4}$   &  83.9  &  33.4  &   16.1 & \textbf{15.6} & 24.7 & 17.3  &   19.5 &  16.7  &  19.1  & \textbf{6.2}  &  6.4 \\
    $\text{ECOC}_{16,2}$   &  84.0  &  41.5  & 23.0 & \textbf{22.7} & 31.0 & 24.5  & 29.0  &  24.5  &  28.7  & \textbf{11.1}  & 11.9  \\
    $\text{ECOC}_{16,1}$   &  85.2  &  49.0  &  \textbf{29.0} & \textbf{29.0}  & 33.4 &  30.7  &  32.2  &  30.7  &  31.9  &  18.7  &  \textbf{15.8}  \\\midrule
    $\text{ECOC}_{32,4}$   &  83.4  &  47.0  &  21.5 & 21.3 & 30.3 &  21.1  & 22.1  & \textbf{20.9}   & 21.8  &  5.5  &  \textbf{4.4} \\
    $\text{ECOC}_{32,2}$   &  83.4 &  52.1  & 26.9 & \textbf{26.8} & 33.2  &   27.5  & 28.6  & 27.4  &  28.7  & 12.0  & \textbf{10.9}  \\
    $\text{ECOC}_{32,1}$   &  85.0  &  57.6  &  \textbf{28.5} & \textbf{28.5} & 36.6 &  29.7  &  31.7  &  29.4  &    31.8 &  15.8  &  \textbf{13.8} \\
    \bottomrule
  \end{tabular}
\end{sc}
\end{small}
\end{center}
\vskip -0.1in
\end{table*}

The results in tables \ref{tab:adv attacks} and \ref{tab:adv attacks fmnist} show the superiority of the ECOC models over the baseline models. For instance, the most robust model in Table~\ref{tab:adv attacks} is $\text{ECOC}_{16,1}$. It has an accuracy of 26.4\% and 27.0\% against the strongest $\ell_{\infty}$ and $\ell_{2}$ attacks, respectively, while the three baseline models all have an accuracy of almost 0.0\% against at least one $\ell_{\infty}$ and $\ell_{2}$ attack. However, we note better robustness gains for ECOC models trained on CIFAR10 (Tab.~\ref{tab:adv attacks}) compared to Fashion-MNIST (Tab.~\ref{tab:adv attacks fmnist}). We believe that ECOC deals better with complex datasets, as the binary classifiers are able to learn more diverse representations leading to better robustness. We leave this point for future investigations.

\begin{figure}[t]
\vskip 0.2in
\begin{center}
\includegraphics[width=0.99\columnwidth]{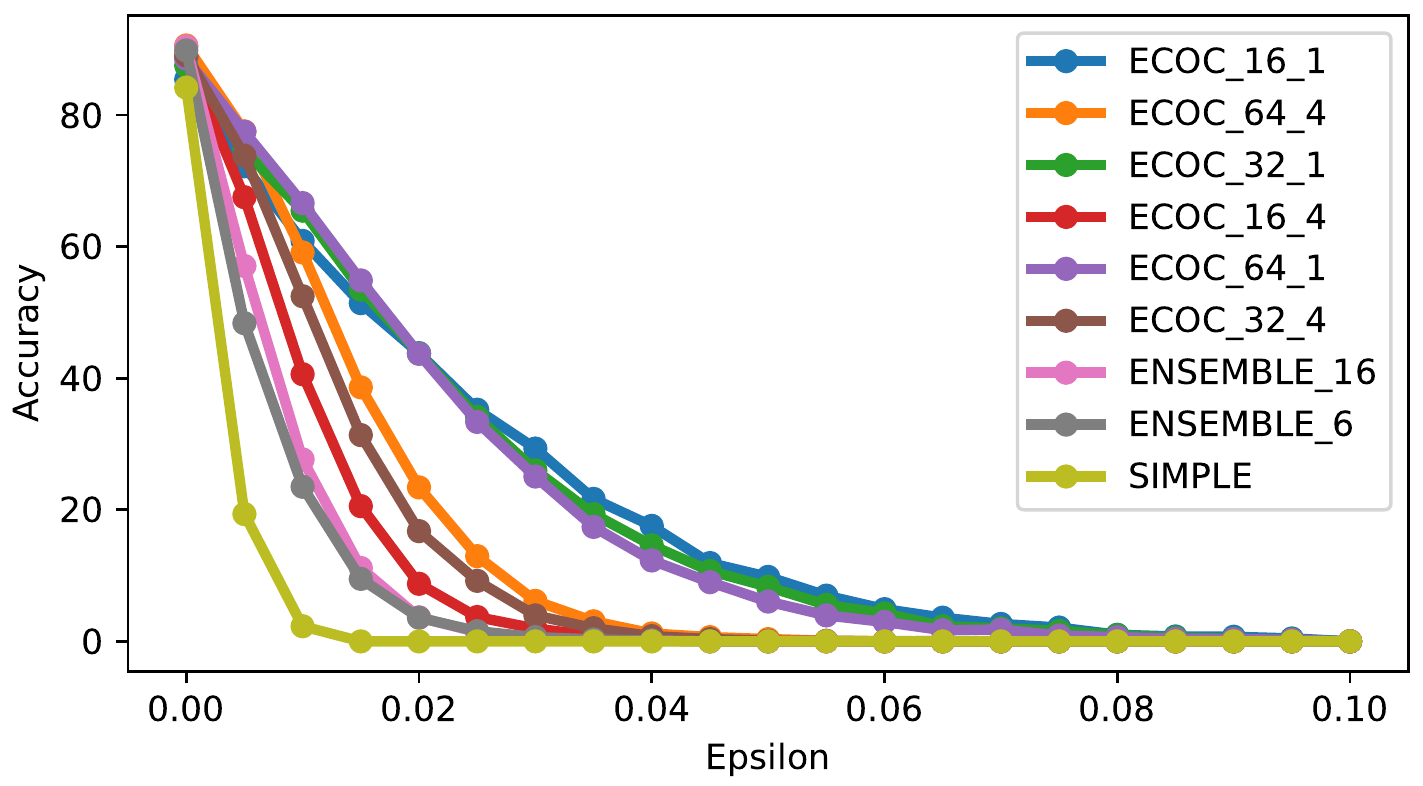}
\caption{Accuracy of ECOC ensembles trained on CIFAR10 against images perturbed with $\text{PGD}_{cw}^{es}$ when varying the $\ell_{\infty}$ bound (Epsilon) of the attack.}
\label{fig:epsilon plot}
\end{center}
\vskip -0.2in
\end{figure}

Considering the results for ECOC ensembles in Table~\ref{tab:adv attacks}, we notice that the robustness to both $\ell_{\infty}$ and $\ell_{2}$ attacks increases as the level of feature extractor sharing decreases. For instance, $\text{ECOC}_{16,4}$, $\text{ECOC}_{16,2}$ and $\text{ECOC}_{16,1}$ have an accuracy of 1.4\%, 7.1\% and 26.4\%, respectively, against the $\text{PGD}^{es}$ attack. Table~\ref{tab:adv attacks fmnist} shows similar results for the 16- and 32-bit ECOC models, supporting the claim that ECOC ensembles with fully independent base classifiers achieve better robustness.


To verify that gradient masking is not the cause of increased robustness, Fig.~\ref{fig:epsilon plot} presents the accuracy of the models from Table~\ref{tab:adv attacks} with respect to the $\ell_{\infty}$ norm of the $\text{PGD}_{cw}^{es}$ perturbations. Since the accuracy is approximately 0\% at an $\epsilon$ value of 0.08 for all models, we conclude that gradient masking is not occurring or at least has a very limited effect \cite{athalye2018obfuscated}.

\subsection{Ensemble Diversity and Robustness} 

\begin{figure}[t]
\centering     
\subfigure[$\text{ECOC}_{16,4}$]{\label{fig:a}\includegraphics[width=0.49\columnwidth]{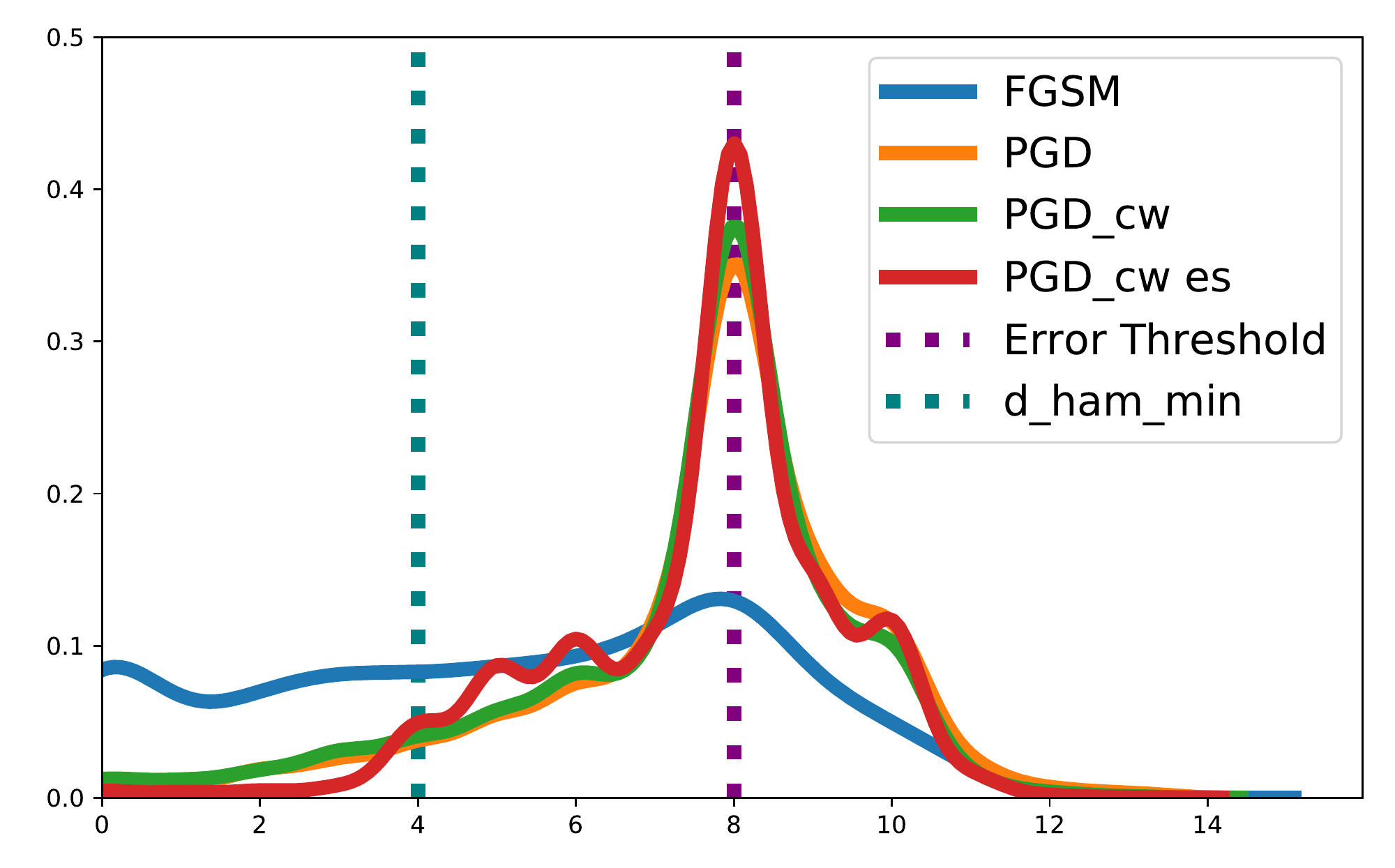}}
\subfigure[$\text{ECOC}_{16,1}$]{\label{fig:b}\includegraphics[width=0.49\columnwidth]{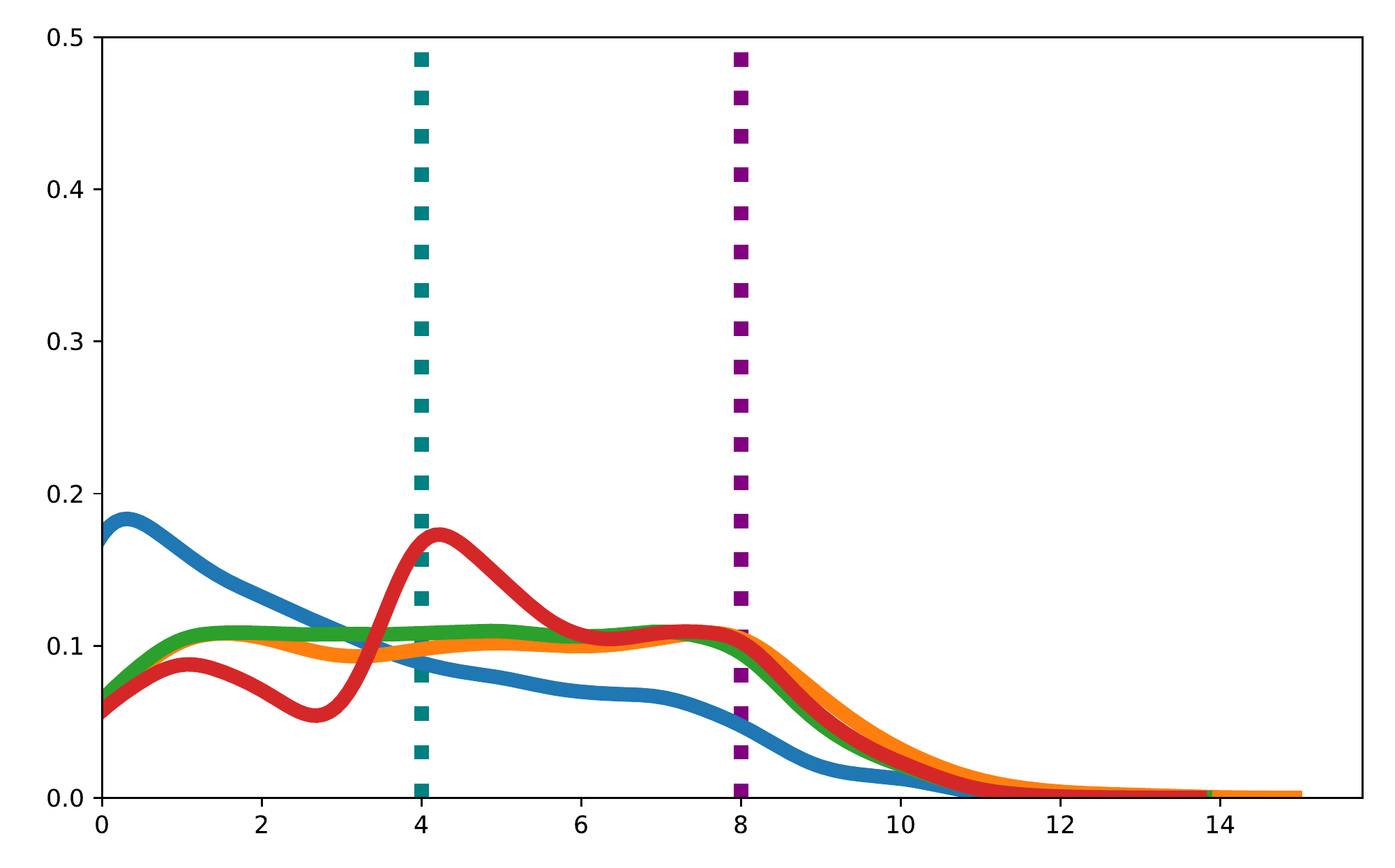}}
\subfigure[$\text{ECOC}_{32,4}$]{\label{fig:c}\includegraphics[width=0.49\columnwidth]{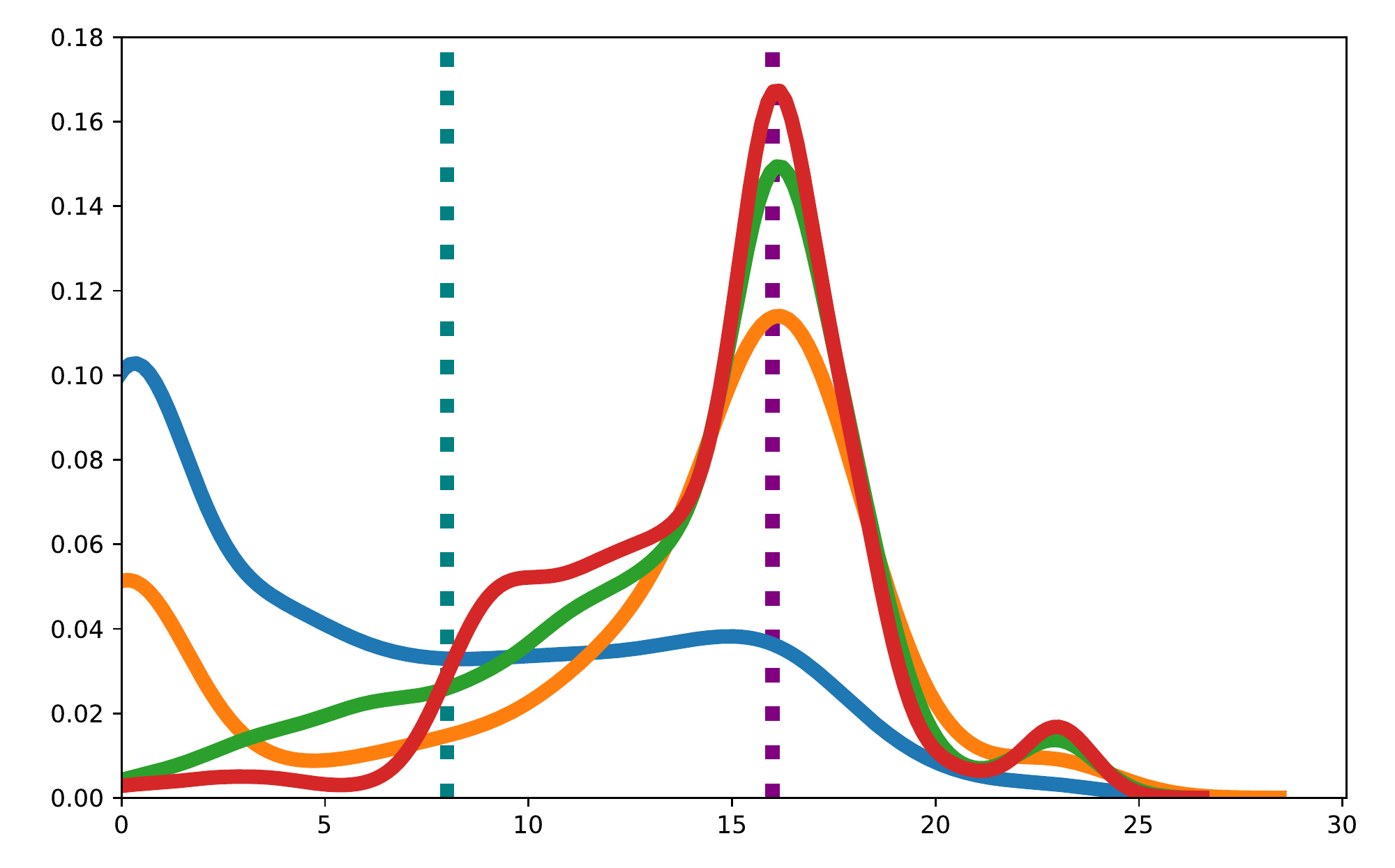}}
\subfigure[$\text{ECOC}_{32,1}$]{\label{fig:d}\includegraphics[width=0.49\columnwidth]{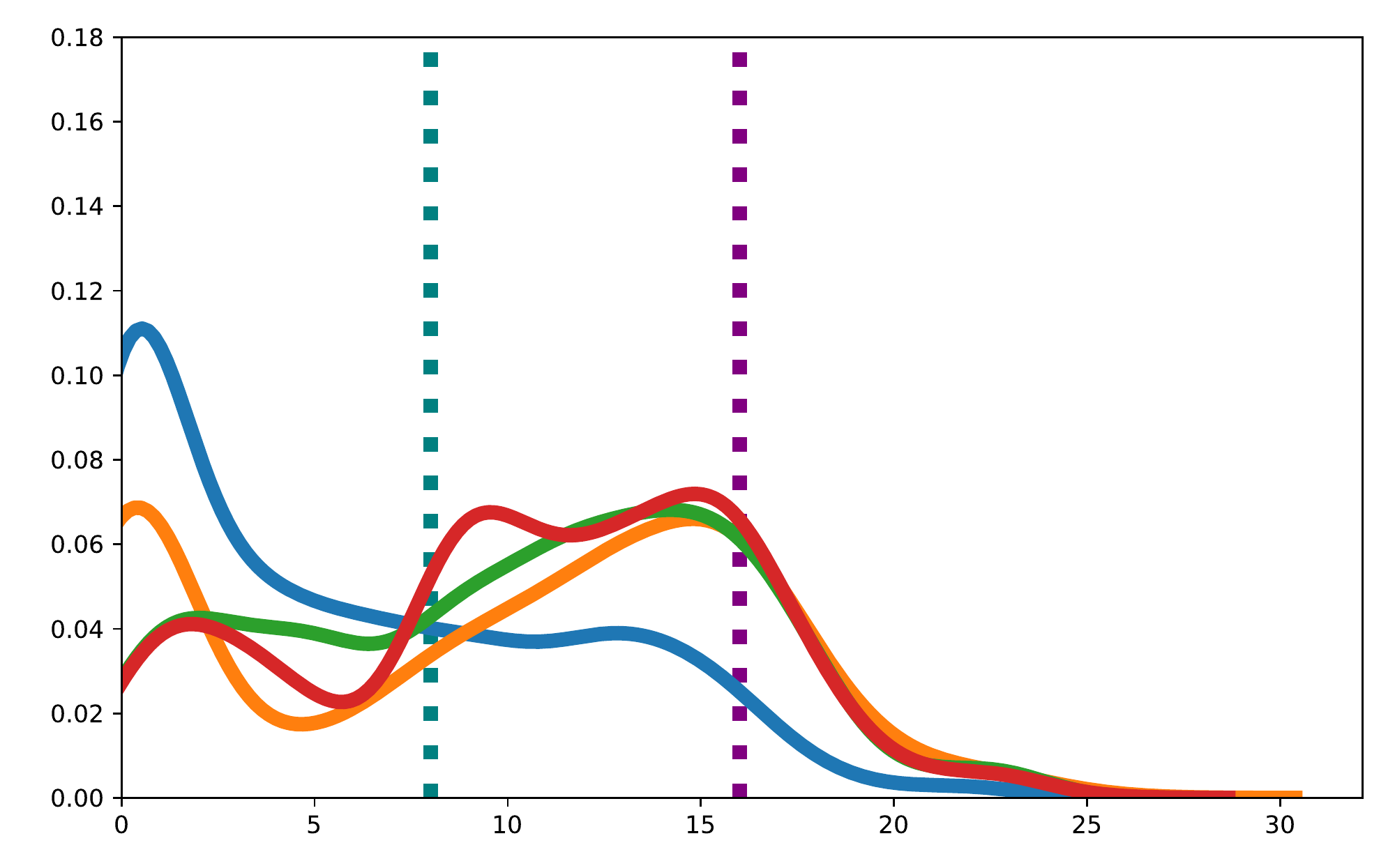}}
\subfigure[$\text{ECOC}_{64,4}$]{\label{fig:e}\includegraphics[width=0.49\columnwidth]{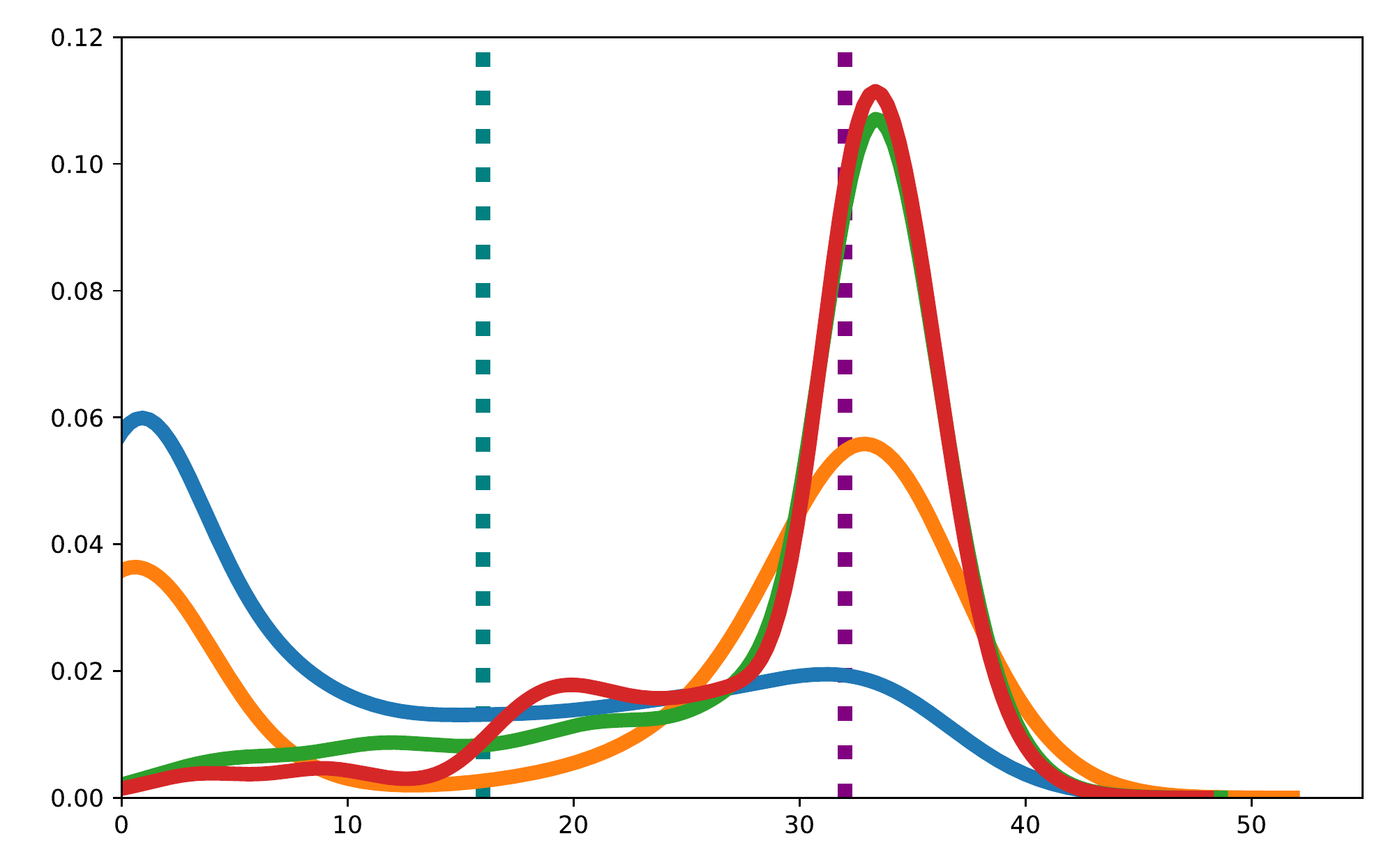}}
\subfigure[$\text{ECOC}_{64,1}$]{\label{fig:f}\includegraphics[width=0.49\columnwidth]{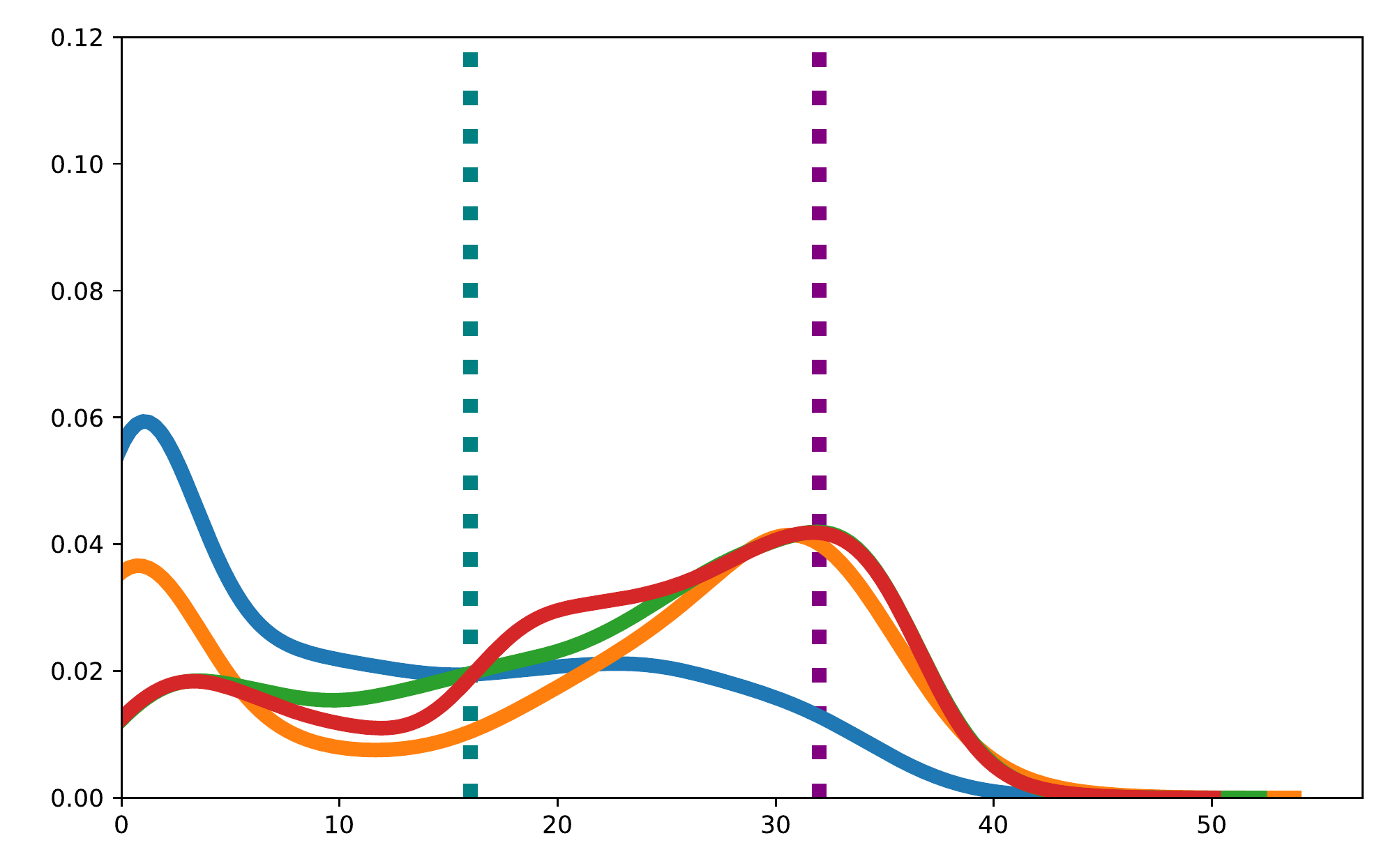}}
\caption{Distribution of the Hamming distance between the predicted and the true codewords for different ECOC ensembles trained on CIFAR10 for various adversarial attacks. The left vertical line represents the minimum Hamming distance of $\lfloor\frac{\theta_\text{minham}-1}{2}+1\rfloor$ required to fool the ECOC ensembles, or error threshold. The right vertical line represents $\theta_\text{minham}$ for each configuration.}
\label{fig:kernel}
\end{figure}

Fig.~\ref{fig:kernel} shows, for ECOC models trained on CIFAR10, distributions of Hamming distances between predicted and true codewords for images perturbed with different adversarial attacks. We argue that this is a good indicator of diversity among the binary classifiers. With more diverse ECOC ensembles, less binary classifiers are fooled simultaneously by adversarial perturbations. For instance, models with shared feature extractors (Fig.~\ref{fig:a}, \ref{fig:c} and \ref{fig:e}) peak at $\theta_\text{minham}$, indicating that the number of classifiers fooled is maximized. However, for ECOC models with independent binary classifiers (Fig.~\ref{fig:b}, \ref{fig:d} and \ref{fig:f}), the distribution is more spread toward the error threshold line, indicating that less binary classifiers are fooled and suggesting better ensemble diversity. Sec.~\ref{app:allKernels} of the supplementary material presents these results for more models.

\subsection{Codeword Length}
ECOC ensembles can correct up to $\lfloor\frac{\theta_\text{minham}-1}{2}\rfloor$ misclassifications from the binary classifiers. Since longer codewords have larger $\theta_\text{minham}$ values, ECOC ensembles with 32- and 64-bit codewords can correct more errors than 16-bit codewords. However, we observe in Table~\ref{tab:adv attacks} and \ref{tab:adv attacks fmnist} that the most robust ECOC model is $\text{ECOC}_{16,1}$. We believe that this is due to better row separation since $\theta_\text{div}$ and $\theta_\text{cdiv}$ are maximized for 16-bit codewords since it is more challenging to maximize row separation with longer codewords. Since better row separation leads to more diverse binary classifiers, it could also lead to better robustness. Results in Fig.~\ref{fig:b}, \ref{fig:d} and \ref{fig:f} confirm this claim, with distribution for $\text{ECOC}_{16,1}$ being much flatter with a small peak around the error threshold line and no peak at $\theta_\text{minham}$. This shows that maximizing the row separation of the codeword matrix to promote ensemble diversity is important for the robustness of ECOC ensembles. 

\begin{table}[t]
\caption{Accuracy against adversarial attacks of an $\text{ECOC}_{16, 1}$ model trained on CIFAR10 with IndAdvT and RegAdvT. PGD iter is the number of PGD iterations used for generating the adversarial perturbations from the batch images.}
\label{tab:adversarial training}
\vskip 0.1in
\begin{center}
\begin{small}
\setlength{\tabcolsep}{3pt}
\begin{sc}
 \begin{tabular}{ccccccc}
    \toprule
    Training Method  &  PGD iter & Clean &  $\text{PGD}^{es}$ & $\text{PGD}_{cw}^{es}$  \\
    \midrule
    IndAdvT  & 2 & 80.7   &  41.6   &  41.7        \\ 
    IndAdvT  & 5 &  75.2  &  44.2   &   44.6     \\ \midrule
    RegAdvT & 2 &  80.5  &  34.2  & 35.4  \\
    RegAdvT  & 5 &  78.2   &   36.4   & 38.1     \\
    RegAdvT  & 10 & 77.5   &  34.8    &  36.1    \\
    \bottomrule
  \end{tabular}
\end{sc}
\end{small}
\end{center}
\vskip -0.1in
\end{table}
\subsection{Adversarial Training}
The results for the last experiment are shown in Table \ref{tab:adversarial training}, where an $\text{ECOC}_{16, 1}$ model is trained on CIFAR10 with adversarial training instances generated for the individual classifiers (IndAdvT) and for the complete ECOC ensemble (RegAdvT). Training and attack parameters are similar to those used previously for the ensembles trained on CIFAR10. Table \ref{tab:adversarial training} specifies the number of PGD iterations used while the step size is $\frac{\epsilon}{3}$ with $\epsilon = 0.031$. 

IndAdvT model with 2 PGD iterations achieved the best robustness-accuracy tradeoff, with a test accuracy of 80.7\% and an accuracy of 41.6\% against the $\text{PGD}^{es}$ attack. Compared to the $\text{ECOC}_{16, 1}$ model from Table~\ref{tab:adv attacks}, which achieved the best results among models trained without adversarial training, this is a 15.2\% robustness improvement with only a 5.1\% test accuracy drop. This shows the benefit of using IndAdvT to improve the robustness of ECOC ensembles. 

The superiority of IndAdvT training over RegAdvT is also illustrated in Table~\ref{tab:adversarial training}. Results with RegAdvT and 2 PGD iterations show a similar test accuracy to IndAdvT, but only an accuracy of 34.2\% against the $\text{PGD}^{es}$ attack (a 7.4\% robustness drop compared to IndAdvT). Results for $\text{ECOC}_{16, 1}$ with RegAdvT also show that the accuracy against $\text{PGD}^{es}$ is limited to about 35-36\% even when more PGD iterations are used to generate the perturbations. This supports our claim that training with perturbations generated on the whole ensemble does not allow the binary classifiers to learn features that are as robust as those learned with classifier-specific perturbations.

\section{Conclusion}

The proposed method and several experiments on robustness of ECOC ensembles against adaptive attacks have demonstrated that relying on independent base classifiers and preserving good row separation in the codeword matrix make ECOC ensembles more robust to adversarial attacks. We also demonstrated that adversarial training of ECOC ensembles works particularly well with perturbations generated on the binary outputs of each individual classifier and not on the whole ensemble.

\section*{Acknowledgements}
This work was supported by funding from NSERC-Canada and CIFAR.

\bibliography{ecoc_paper}
\bibliographystyle{icml2022}

\newpage
\appendix
\onecolumn

\section{CNN Architecture}
\label{app:architecture}
\begin{table*}[b]
\caption{Base architecture of the neural networks used in this work. Parameters A, B, C and D correspond respectively to 32, 64, 128 and 16 for models trained on CIFAR10, and 32, 32, 32, and 4 for models trained on Fashion-MNIST.}
\label{tab:cnn architecture}
\vskip 0.1in
\begin{center}
\begin{small}
\begin{sc}
 \begin{tabular}{lcccc}
    \toprule
      Layer & Number of Filters & Kernel Size & Stride & Padding \\
      \midrule
      Conv 1 + ReLu   &  A  &  5 $\times$ 5 &   1  &  2\\ 
      Conv 2 + ReLu   &  A  &  5 $\times$ 5 &   1  &  2\\
      Conv 3 + ReLu   &  A  &  3 $\times$ 3 &   2  &  1\\
      Conv 4 + ReLu   &  B  &  3 $\times$ 3 &   1  &  1\\
      Conv 5 + ReLu   &  B  &  3 $\times$ 3 &   1  &  1\\
      Conv 6 + ReLu   &  B  &  3 $\times$ 3 &   2  &  1\\
      Conv 7 + ReLu   &  C  &  3 $\times$ 3 &   1  &  1\\
      Conv 8 + ReLu   &  C  &  3 $\times$ 3 &   1  &  1\\
      Conv 9 + ReLu   &  C  &  3 $\times$ 3 &   2  &  1\\
      Conv 10 + ReLu  &  D  &  2 $\times$ 2 &   1  &  1\\
      Conv 11 + ReLu  &  D  &  2 $\times$ 2 &   1  &  0\\
      Fully Connected & --  & --            & --  & --\\
    \bottomrule
  \end{tabular}
\end{sc}
\end{small}
\end{center}
\vskip -0.1in
\end{table*}

Table \ref{tab:cnn architecture} presents the base CNN architecture used in this work. It is inspired by the architecture used by \citet{vermaECOC}. No batch normalization layers are used as this has been shown to increase adversarial vulnerability through the learning of non-robust features \cite{Benz_2021_ICCV}.

For ECOC ensembles with independent binary classifiers, each classifier consists of the 11 convolution layers and the fully connected layer from Table~\ref{tab:cnn architecture}. The fully connected layer is composed of a single neuron. For ECOC ensembles with shared feature extractors, each shared feature extractor consists of the first 9 convolution layers in Table~\ref{tab:cnn architecture}. The parallel heads connected to each shared feature extractor consist of independent copies of the last two convolution layers and the fully connected layer from Table~\ref{tab:cnn architecture}. The fully connected layer is again composed of a single neuron. For the baseline models, each $\text{SIMPLE}$ model is composed of all the convolution layers followed by the fully connected layer of Table~\ref{tab:cnn architecture}, with the fully connected layer outputting 10 values since the cross-entropy loss is used.

\section{Adversarial Attack Selection} 
\label{app:adv attacks}
Three types of white-box attacks have been used in our experiments:
\begin{itemize}
\item FGSM \cite{goodfellow2014} is selected because it is a common attack that is fast and convenient to perform at first. It is a good indicator of potential problems with more powerful attacks since the accuracy against FGSM should always be better than the accuracy against more powerful attacks.
\item PGD \cite{madry2017towards} is selected as it is one of the most powerful $\ell_\infty$ attacks as noted by \citet{carlini2019evaluating}, while also being quite common.
\item $\text{C\&W}_{\ell_2}$ was selected since it is one of the most powerful $\ell_2$ attacks available \cite{carlini2019evaluating}. This attack is unbounded, generating the smallest $\ell_2$ perturbations possible causing misclassifications. A threshold is set to manually bound the attack and discard perturbations of greater distortions. It allows the robustness against this attack to be evaluated with an accuracy rate, the same way the robustness against the other bounded attacks is evaluated \cite{carlini2019evaluating}. The $\ell_2$ threshold is set to $1.0$ for CIFAR10 and $1.5$ for Fashion-MNIST. 
\end{itemize}

\section{Kernel Distributions of All ECOC Models}
\label{app:allKernels}
\begin{figure}[t]
\centering     
\subfigure[$\text{ECOC}_{16,4}$]{\label{fig:g}\includegraphics[width=0.3\textwidth]{figures/ecoc_4_4.pdf}}
\subfigure[$\text{ECOC}_{16,2}$]{\label{fig:h}\includegraphics[width=0.3\textwidth]{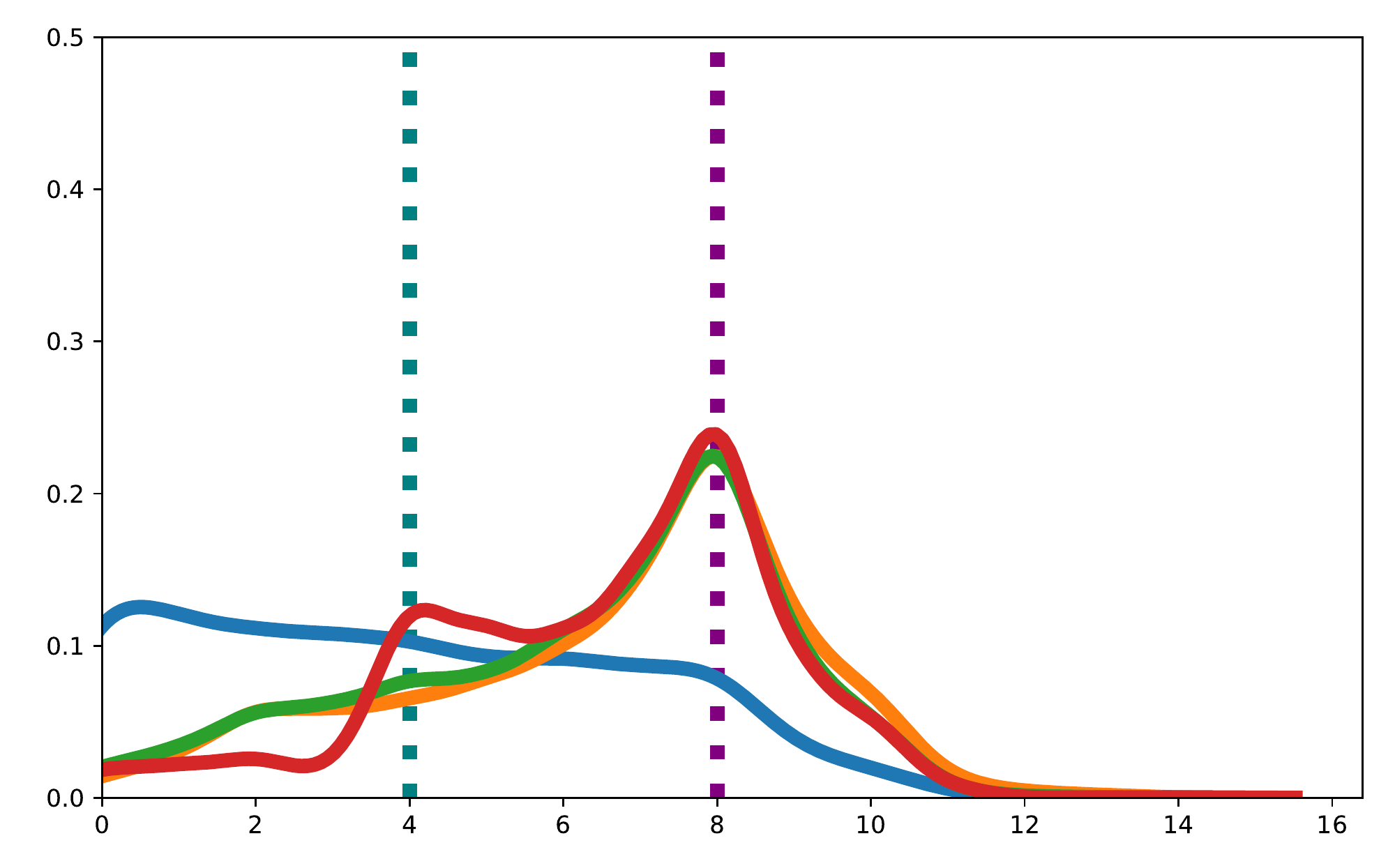}}
\subfigure[$\text{ECOC}_{16,1}$]{\label{fig:i}\includegraphics[width=0.3\textwidth]{figures/ecoc_16_1.pdf}}
\subfigure[$\text{ECOC}_{32,4}$]{\label{fig:j}\includegraphics[width=0.3\textwidth]{figures/ecoc_8_4.pdf}}
\subfigure[$\text{ECOC}_{32,2}$]{\label{fig:k}\includegraphics[width=0.3\textwidth]{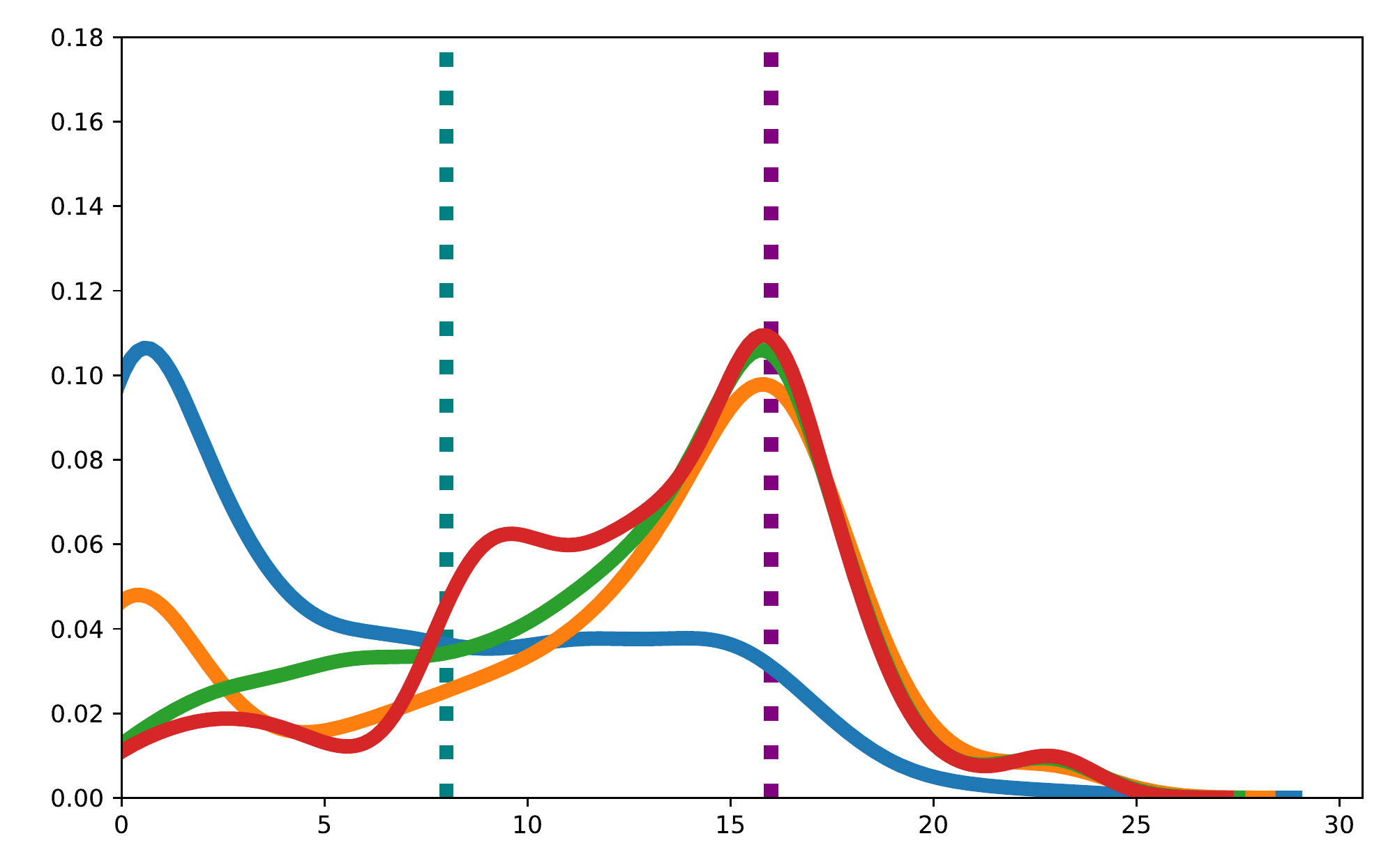}}
\subfigure[$\text{ECOC}_{32,1}$]{\label{fig:l}\includegraphics[width=0.3\textwidth]{figures/ecoc_32_1.pdf}}
\subfigure[$\text{ECOC}_{64,4}$]{\label{fig:m}\includegraphics[width=0.3\textwidth]{figures/ecoc_16_4.pdf}}
\subfigure[$\text{ECOC}_{64,2}$]{\label{fig:n}\includegraphics[width=0.3\textwidth]{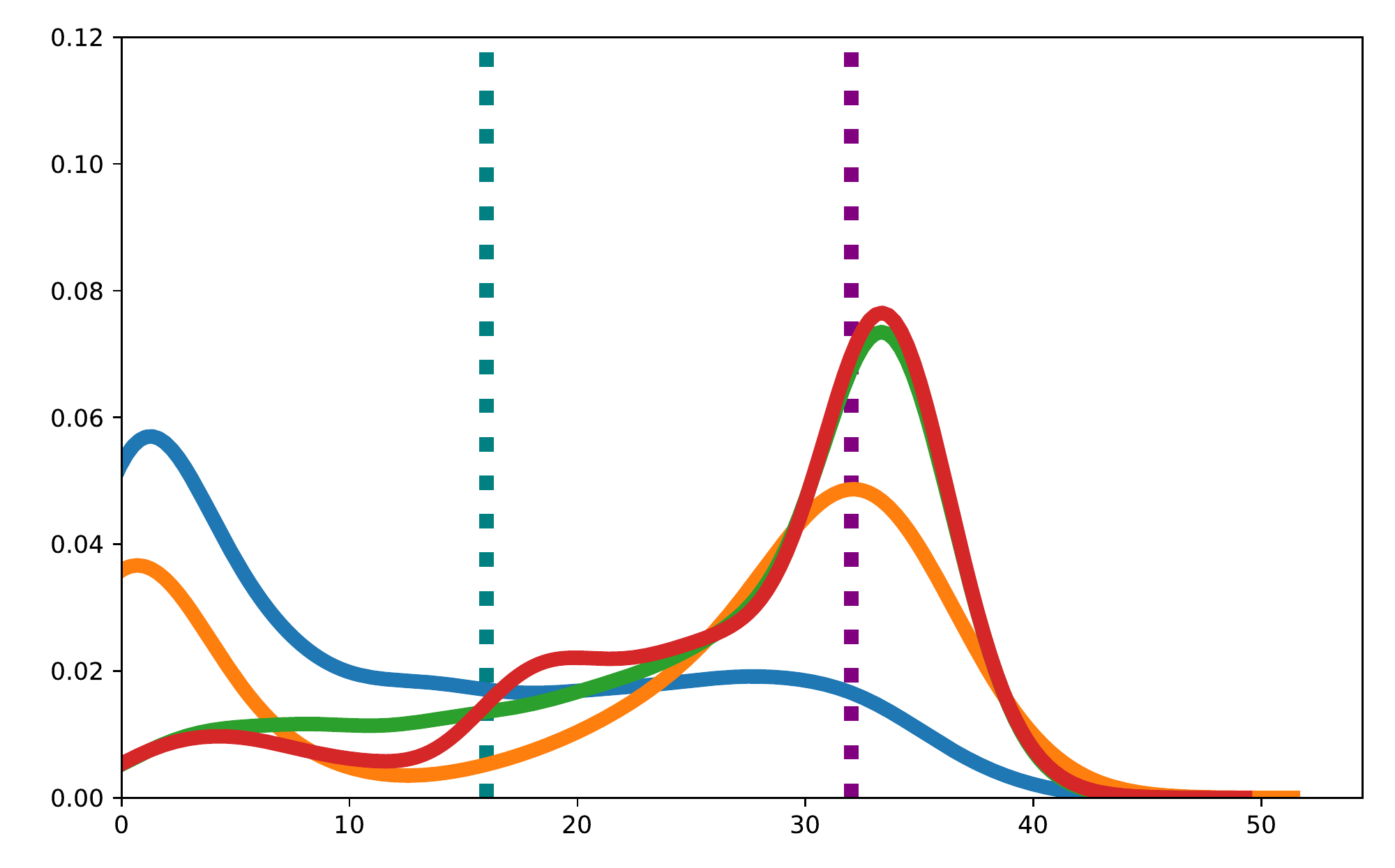}}
\subfigure[$\text{ECOC}_{64,1}$]{\label{fig:o}\includegraphics[width=0.3\textwidth]{figures/ecoc_64_1.pdf}}
\caption{Distribution of the hamming distance between the predicted and the true codewords for all ECOC ensembles trained on CIFAR10 from Table \ref{tab:adv attacks} against images perturbed with different adversarial attacks.}
\label{fig:kernel all}
\end{figure}
Fig.~\ref{fig:kernel all} is similar to Fig.~ \ref{fig:kernel} with the addition of the distributions for $\text{ECOC}_{16,2}$, $\text{ECOC}_{32,2}$ and $\text{ECOC}_{64,2}$. This indicates that for the same number of bits in the codewords, the more independent the  ECOC binary classifiers are (i.e.~less feature extractors are shared), the less binary classifiers are fooled simultaneously by the perturbations.

\end{document}